\documentclass[10pt, a4paper]{article}

\usepackage{lrec-coling2024} 

\usepackage{bm}
\usepackage{amsmath,amssymb}
\usepackage{bbm}
\usepackage{multirow}
\usepackage{makecell}
\usepackage{booktabs}
\usepackage{pifont}
\usepackage{subcaption}
\usepackage{arydshln}
\usepackage{graphicx}
\usepackage{CJKutf8}

\title{Building a Japanese Document-Level Relation Extraction Dataset Assisted by Cross-Lingual Transfer}

\name{Youmi Ma, An Wang, Naoaki Okazaki} 

\address{Tokyo Institute of Technology \\
         Tokyo, Japan\\
         \{youmi.ma@nlp., an.wang@nlp., okazaki@\}c.titech.ac.jp\\}

\abstract{
Document-level Relation Extraction (DocRE) is the task of extracting all semantic relationships from a document. While studies have been conducted on English DocRE, limited attention has been given to DocRE in non-English languages. This work delves into effectively utilizing existing English resources to promote DocRE studies in non-English languages, with Japanese as the representative case. As an initial attempt, we construct a dataset by transferring an English dataset to Japanese. However, models trained on such a dataset suffer from low recalls. We investigate the error cases and attribute the failure to different surface structures and semantics of documents translated from English and those written by native speakers. We thus switch to explore if the transferred dataset can assist human annotation on Japanese documents. In our proposal, annotators edit relation predictions from a model trained on the transferred dataset. Quantitative analysis shows that relation recommendations suggested by the model help reduce approximately 50\% of the human edit steps compared with the previous approach. Experiments quantify the performance of existing DocRE models on our collected dataset, portraying the challenges of Japanese and cross-lingual DocRE.
 \\ \newline  }

\begin{document}

\maketitleabstract

\section{Introduction}

Document-level Relation Extraction (DocRE) aims to identify all semantic relationships between entities in a document~\cite{yao-etal-2019-docred}.
The task promotes Relation Extraction (RE) to a more practical setting, where relations can reside between entity pairs \textit{document-wise}, i.e., within and beyond the sentence boundary.
DocRE is worth spotlighting as it not only inherits the significance of RE in benefiting knowledge graph completion and question answering but also showcases how models comprehend long text~\cite{yu-etal-2017-improved, trisedya-etal-2019-neural, chen-etal-2023-models}.
Even in the era of large language models (LLMs), the task deserves more attention as in-context learning of DocRE was considered not yet feasible~\cite{wadhwa-etal-2023-revisiting}. 

DocRE research has been conducted mainly in English~\cite{yao-etal-2019-docred,zhou2021atlop,tan-etal-2022-revisiting}.
This work aims to promote DocRE in other languages with the help of English resources.
Specifically, we utilize existing resources of English DocRE to construct datasets and models for non-English DocRE.
We chose Japanese as our target language for the following two reasons.
Firstly, despite Japanese being a widely used language for web content, there is currently a notable absence of general-purpose Japanese DocRE resources.
Our work thus contributes to the community by establishing the foundation for Japanese DocRE.
Secondly, Japanese stands out as one of the most linguistically distant languages from English~\cite{RePEc:iza:izadps:dp1246}. 
The dissimilarity encompasses various aspects, including script models and word order. 
Therefore, our research setting is highly representative, and the insights we gain will hold value when acquiring resources for other languages.

\begin{figure}[!t]
    \centering
    \includegraphics[width=.49\textwidth]{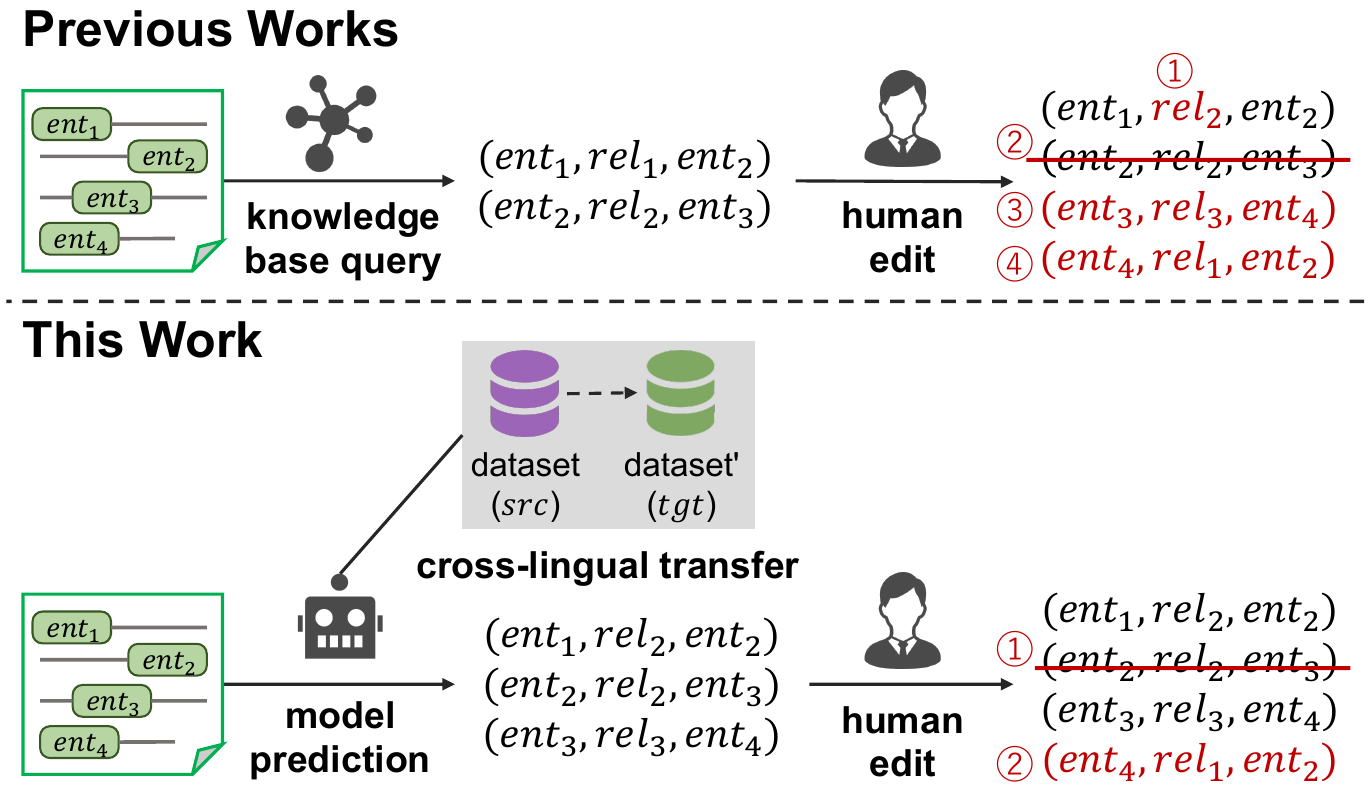}
    \caption{Overview of the proposed annotation scheme. \textit{src} and \textit{tgt} represent the source and target language, respectively. Previous works require 4 human edit steps to reach the final annotation, while ours only require 2.}
    \label{fig:overview}
\end{figure}

\begin{table*}[!t]
    \centering
    \small
    \begin{tabular}{lcccccc}
        \Xhline{3\arrayrulewidth}
        \textbf{Dataset} & \textbf{Lang.} & \textbf{\# Triples } & \textbf{\# Docs.} & 
         \textbf{Avg. \# Toks.} & \textbf{\# Rels.} & \textbf{Evi.} \\
        \Xhline{2\arrayrulewidth}
        DocRED~\cite{yao-etal-2019-docred} & \textit{en.} & 50,503 & 4,051 &  198.4 & 96 & Y \\
        Re-DocRED~\cite{tan-etal-2022-revisiting} & \textit{en.} & 120,664 & 4,053 & 198.4 & 96 & N \\
        HacRED~\cite{cheng-etal-2021-hacred} & \textit{zh.} & 56,798 & 7,731 & 122.6 & 26 & N \\
        HistRED~\cite{yang-etal-2023-histred} & \textit{kr.} & 9,965 & 5,816 & 100.6 & 20 & Y \\        
        \hline
        \textbf{JacRED} (Ours) & \textit{ja.} & 42,241 & 2,000 & 260.1 & 35 & Y \\
        \Xhline{3\arrayrulewidth}
    \end{tabular}
    \caption{Statistics of existing and proposed DocRE datasets. Column \textbf{Evi.} shows whether each dataset annotates evidence sentences or not. Statistics for DocRED are from the human-annotated subset.}
    \label{tab:datasets}
\end{table*}

We first explore if DocRE resources of high quality can be obtained with zero human effort.
To this end, we automatically construct a Japanese DocRE dataset with cross-lingual transfer.
Specifically, we translate Re-DocRED~\cite{tan-etal-2022-revisiting}, a popular English DocRE dataset of high quality, into Japanese with a machine translator.
An automatically constructed dataset (hereafter referred to as \textit{Re-DocRED\textsuperscript{ja}}) can thus be obtained without human annotators.
The translation-based cross-lingual transfer has been successfully applied to other information extraction (IE) tasks, including named entity recognition and sentence-level relation extraction~\cite{chen-etal-2023-frustratingly,hennig-etal-2023-multitacred}.
However, we observe that models trained on Re-DocRED\textsuperscript{ja} suffer from low recalls when extracting relation triples from raw Japanese text.
We investigate the error cases and attribute the failures to the discrepancies between documents in Re-DocRED\textsuperscript{ja} and those composed by native speakers.
The discrepancies include deviations of topics and wording. 
Our observation underscores the uniqueness and complexity of DocRE in comparison to other IE tasks.

Given that Re-DocRED\textsuperscript{ja} is not suitable for immediate practical application, we explore if the dataset can assist human annotation.
As in Figure~\ref{fig:overview}, we adopt a semi-automatic, edit-based annotation scheme, where annotators edit machine recommendations by removing incorrect instances and supplementing missed instances~\cite{yao-etal-2019-docred,cheng-etal-2021-hacred,tan-etal-2022-revisiting}.
In contrast to previous works where only relation instances from an existing knowledge base are recommended~\cite{yao-etal-2019-docred,cheng-etal-2021-hacred}, we recommend instances with a state-of-the-art DocRE model trained on Re-DocRED\textsuperscript{ja}.
The collected dataset is named as \textbf{JacRED} (\underline{Ja}panese Do\underline{c}ument-level \underline{R}elation \underline{E}xtraction \underline{D}ataset), with statistics shown in Table~\ref{tab:datasets}.
We quantitatively analyze recommendations from the model trained on Re-DocRED\textsuperscript{ja} and those from knowledge base queries and find the former reduces the human edit steps to half of the latter.

We employ JacRED as a benchmark for evaluation.
Firstly, we evaluate the performance of existing models on Japanese DocRE. 
While models trained using the train set of JacRED perform fairly on the test set, the scores fall short of those achieved on Re-DocRED.
The result indicates that JacRED introduces extra challenges in addition to Re-DocRED.
Notably, we observe that in-context learning of LLMs yields poor performance on JacRED, in line with the findings of~\citet{wadhwa-etal-2023-revisiting}.
Next, we quantify the performance gap between models trained on Re-DocRED\textsuperscript{ja} and those trained on JacRED.
The results further demonstrate that, although translation-based cross-lingual transfer appears effective for a range of IE tasks, it does not hold true for DocRE, especially for distant language pairs.
Additionally, JacRED also enables the evaluation of cross-lingual DocRE.
We assess the cross-lingual transferability of existing DocRE models between English and Japanese, from which we observe challenges due to the complexity of document semantics.
Our dataset will be publicly available\footnote{The dataset is available at \url{https://github.com/YoumiMa/JacRED}}.

\section{Dataset Construction}
\label{sec:dataset}

\paragraph{Task Definition.} 
For each document $D$ consisted of $n$ sentences $\mathcal{X}_D = x_1, x_2, \dots, x_n$, entities within the document are given as $\mathcal{E}_D = \{e_1, e_2, \dots, e_k\}$, where each entity $e_i \in \mathcal{E}_D$ is a collection of all its proper-noun mentions $e_i = \{m_1^i, m_2^i,\dots, m_l^i \}$.
A DocRE model is expected to extract all relation triples within the document in the form of $(e_h, r, e_t)$, where $e_h$ is the head entity, $e_t$ is the tail entity, and $r$ is a relation label chosen from a predefined set. 
Additionally, we also expect the model to perform \textit{evidence retrieval}, where evidence for each relation prediction is provided at the sentence level.
In other words, for a predicted triple $(e_h, r, e_t)$, the model should be able to return the evidence sentences $\mathcal{V}_{{e_h},r,{e_t}} \subseteq \mathcal{X}_D$.

\paragraph{Approach.} We explore ways to construct language resources for Japanese DocRE using existing English resources.
To do so, we start by automatically building a dataset with cross-lingual transfer(Section~\ref{sec:auto}).
The approach has been reported successful in other IE tasks~\cite{chen-etal-2023-frustratingly,hennig-etal-2023-multitacred}; If the transferred dataset portrays the characteristics of Japanese DocRE well, there is no need to recruit human annotators.
However, we observe that the DocRE models trained with such a dataset err on raw Japanese text.
Nevertheless, the model yields predictions of fair quality.
We thus adopt the model trained on the transferred dataset as an intermediary tool to assist human annotation (Section~\ref{sec:manual}).


\subsection{Automatic Construction}
\label{sec:auto}

We build a Japanese version of Re-DocRED~\cite{tan-etal-2022-revisiting}.
Re-DocRED revises DocRED~\cite{yao-etal-2019-docred}, the first and most popular DocRE dataset collected from English Wikipedia.

\begin{figure}[t]
    \centering
    \includegraphics[width=.49\textwidth]{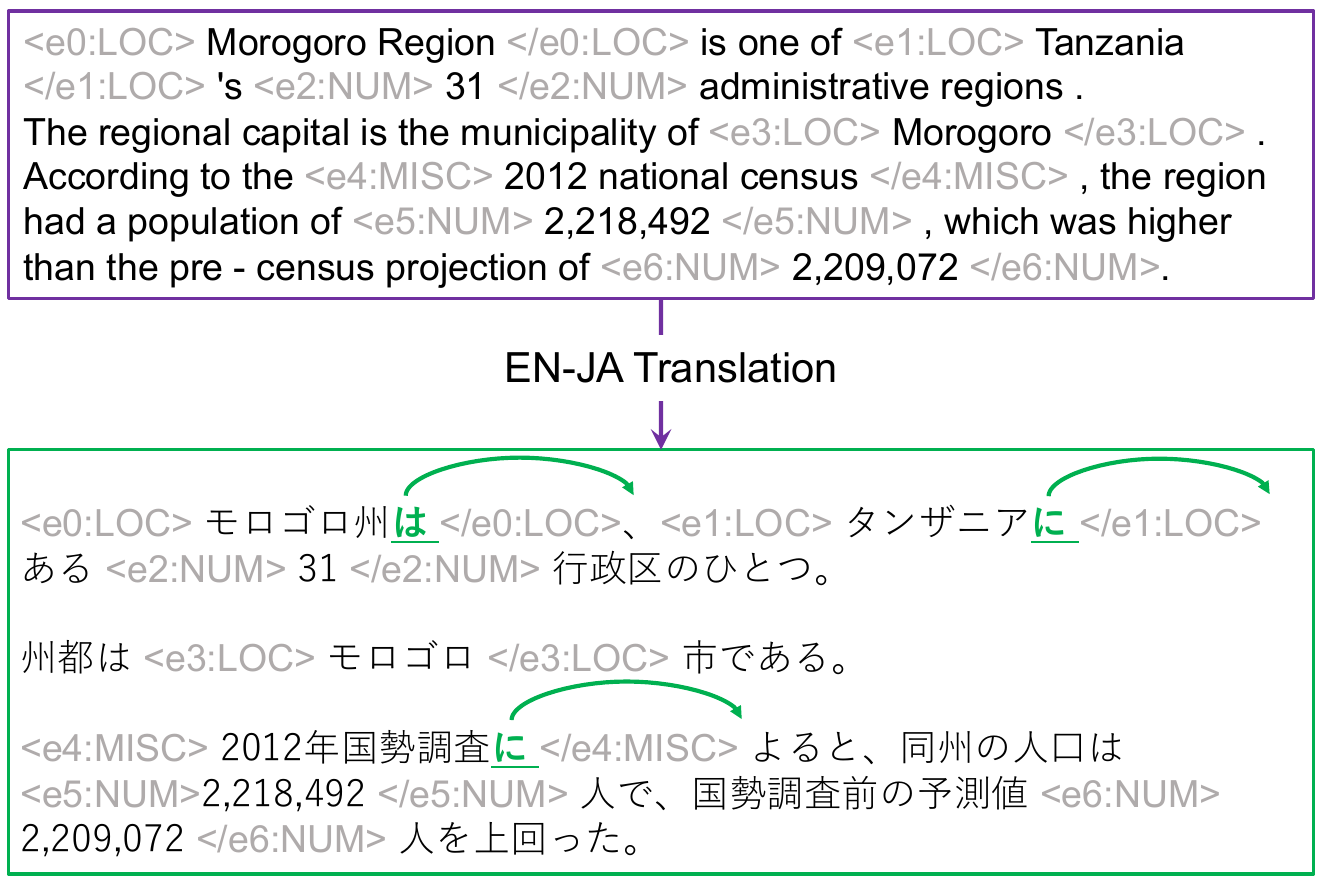}
    \caption{Transferring Re-DocRED from English into Japanese. We post-edit the translation to detach case markers from entity spans.}
    \label{fig:automatic}
\end{figure}

\paragraph{Translation and Annotation Projection.}
We translate the complete train/dev/test splits of Re-DocRED into Japanese with the help of machine translators.
As shown in Figure~\ref{fig:automatic}, XML tags are inserted around each entity.
Documents are translated from English to Japanese with the tags so that entity spans are projected jointly during the translation process.
Relations associated with can be thus directly inherited from the English dataset.
This mark-then-translate method has been reported to work well for multiple structured prediction tasks~\cite{chen-etal-2023-frustratingly}.
We utilize DeepL to perform translation, as it enables translation while preserving XML tag markups\footnote{\url{https://api.deepl.com/v2/translate}}.

\paragraph{Post-processing for Case Markers.}
Given the translation as in Figure~\ref{fig:automatic}, we recognize the necessity of post-editing due to the presence of case markers in entity spans.
Case markers (``kaku-joshi'' in Japanese) are special linguistic units attached to the end of nouns to indicate the relationship between words.
A case marker only reveals the grammatical role but does not contribute to the semantics of the noun phrase it is attached to.
\begin{CJK}{UTF8}{min}
For example, in entity span \textbf{<e0>} of the Japanese translation, a topic marker ``は'' following ``モロゴロ州'' (Morogoro Region) indicates the noun phrase to be the topic of this sentence.

We detach case makers from the entity span with the Japanese morphological analyzer MeCab~\cite{kudo-etal-2004-applying}\footnote{\url{https://taku910.github.io/mecab/}}, removing tokens identified as particles at the end of each span.
The obtained dataset is denoted as Re-DocRED\textsuperscript{ja}.

\begin{figure}[t]
    \centering
    \begin{subfigure}[b]{.48\textwidth}
            \centering
                \includegraphics[width=\textwidth]{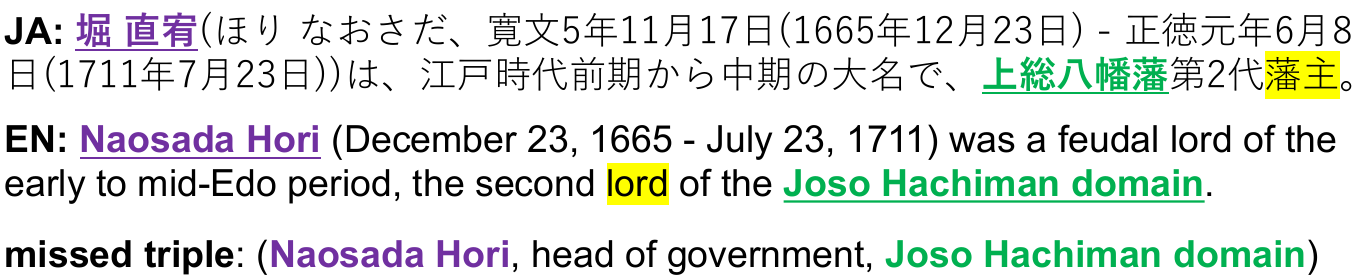}
                \caption{Example of unextracted relations due to the topic shift of contents. The highlighted ``藩主'' is a Japanese historical term used from 1603 to 1912 meaning ``lord''. }
            \label{fig:auto_errs_1}
    \end{subfigure}
    \begin{subfigure}[b]{.48\textwidth}
            \centering
                \includegraphics[width=\textwidth]{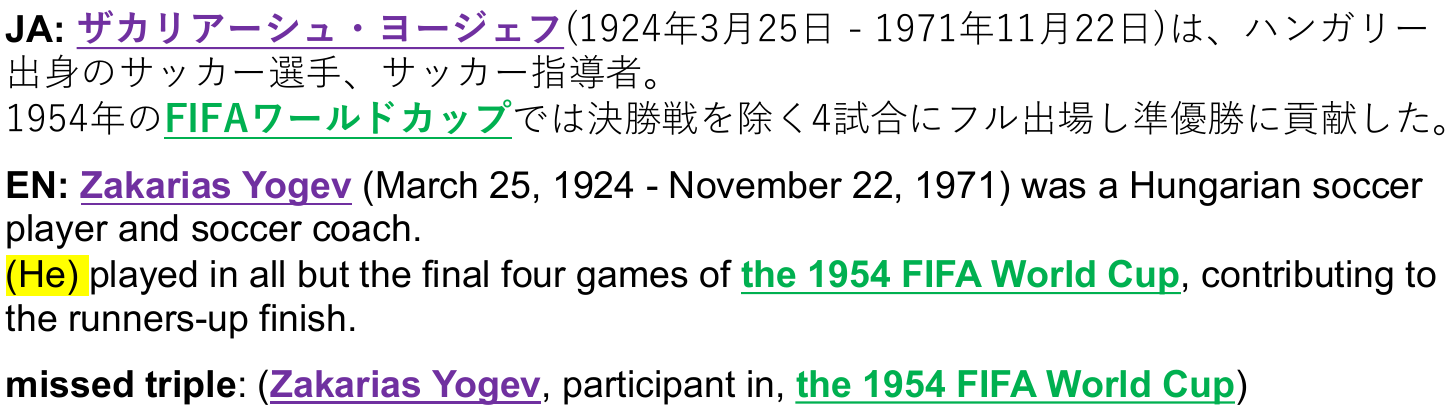}
                \caption{Example of unextracted relations due to the gap of surface structures. The subject of the second sentence is left out in Japanese. }
            \label{fig:auto_errs_2}
    \end{subfigure}        
    \caption{Cases where the model trained on Re-DocRED\textsuperscript{ja} failed to predict. Documents are shown as partial for better visibility. Note that English translations are provided only for reference, while predictions are actually done on Japanese texts.}
    \label{fig:auto_errs}
\end{figure}
\end{CJK}
\paragraph{Limitations of Transferred Dataset.}
When utilizing Re-DocRED\textsuperscript{ja} as the training data and test bed, we witness DREEAM~\cite{ma-etal-2023-dreeam}, the current state-of-the-art DocRE model, achieving an F1 score of 72.74 (cf. the same architecture scores 77.94 on the original Re-DocRED).
However, when ``real'' Japanese documents from Japanese Wikipedia are fed into the model, we observe relation triples being left out in the predictions, with typical examples demonstrated in Figure~\ref{fig:auto_errs}.
Two possible reasons can be raised to explain why the model trained on Re-DocRED\textsuperscript{ja} fails: 
(1) \textbf{Topic Shift of Contents}: Re-DocRED\textsuperscript{ja} cannot represent the real topic distribution of Japanese documents.
Collected from English Wikipedia, Re-DocRED consists of contents that English speakers are concerned about, which do not necessarily match the interests of Japanese speakers.
As in Figure~\ref{fig:auto_errs_1}, Re-DocRED\textsuperscript{ja} lacks documents about Japanese culture, preventing the DocRE models from being localized.
(2) \textbf{Gap of Surface Structures}: The surface structures, i.e., how words are organized in the sentence, of Re-DocRED\textsuperscript{ja} follow the logic of English, which is distinct from that of Japanese.
Figure~\ref{fig:auto_errs_2} showcases a typical example of how Japanese differs from English in surface structures regarding the omission of subjects.
Re-DocRED\textsuperscript{ja} thus cannot reproduce the surface structures of ``real'' Japanese, resulting in failures of the trained model.

\subsection{Semi-Automatic Construction}
\label{sec:manual}

Having observed drawbacks of Re-DocRED\textsuperscript{ja}, we postulate that human annotations are necessary to better depict Japanese DocRE.
We thus involve human annotators in constructing a Japanese DocRE dataset, which we call JacRED.
The annotation process consists of two phases: the entity mention annotation phase and the relation annotation phase.
Both phases follow an edit-based scheme~\cite{yao-etal-2019-docred}: Annotators only need to edit machine recommendations instead of listing all relation instances from scratch.

The quality of machine recommendation is crucial under the edit-based scheme: Poor recommendations require more edits, which will drastically increase the annotators' workload and affect the dataset's quality.
The problem is recognized in DocRED as the \textit{false-negative issue}, where too many relation instances are left out in the recommendations to be mended by human edits~\cite{huang-etal-2022-recommend}.
We propose to mitigate this issue using Re-DocRED\textsuperscript{ja}, utilizing a model trained on Re-DocRED\textsuperscript{ja} to recommend relation instances.

\paragraph{Documents.} 
JacRED is built on top of the Japanese edition of Wikipedia.
We clean up the dump and extract the opening text of each page as the document\footnote{2023-01-01 dump at \url{https://dumps.wikimedia.org/jawiki/}}, with only those longer than 256 characters kept in our annotation pool. 


\paragraph{Annotators.} Given the complexity of the task, we recruit native Japanese speakers with expertise in annotating language resources instead of crowdsourcing\footnote{Measures including the Inter Annotator Agreements (IAA) are thus not reported in this paper.}.
The annotators first work individually on different data and then cross-check the worked annotations.
The annotation tool is BRAT~\cite{stenetorp-etal-2012-brat} during both phases.

\subsubsection{Entity Mention Annotation}

The purpose of the entity annotation phase is two-fold: (1) to obtain high-quality entity mention annotations for each document and (2) to filter out documents involved with few entities and relations.

\paragraph{Entity Types.} We adopt the definition of IREX (Information Retrieval and Extraction
Exercise,~\citet{sekine-isahara-2000-irex}) with 8 types, whose scope is similar to that of DocRED.
A list of entity types is provided in Table~\ref{tab:ent_types} of Appendix A.

\paragraph{Machine Recommendations.}
We parse each document and obtain machine predictions of named entity mentions using KWJA~\cite{ueda-etal-2023-kwja}, a unified analyzer for Japanese.

\paragraph{Document Filtering.}
Another round of document filtering is performed based on the machine prediction to remove documents that are likely to contain few \textbf{cross-sentence} relations.
To this end, we first link each mention to Wikidata entities~\cite{wikidata}.
If an edge with label $r$ connects a certain entity pair $(e_h, e_t)$ in the knowledge base, we treat $(e_h, r, e_t)$ as an extractable relation triple from the document, following the distant-supervision assumption~\cite{mintz-etal-2009-distant}.
Only documents with more than 4 cross-sentence relations are preserved in the annotation pool.
We employ mGENRE~\cite{de-cao-etal-2022-multilingual} for entity linking and KGTK~\cite{ilievski2020kgtk} for connectivity check.

\paragraph{Human Edits.}
We randomly select 2,000 documents from the annotation pool for human annotation.
Human annotators review recommendations in each document, correcting wrongly predicted entity mentions and supplementing missed ones.

\subsubsection{Relation Annotation}
\begin{figure*}[!t]
    \centering
\includegraphics[width=.98\textwidth]{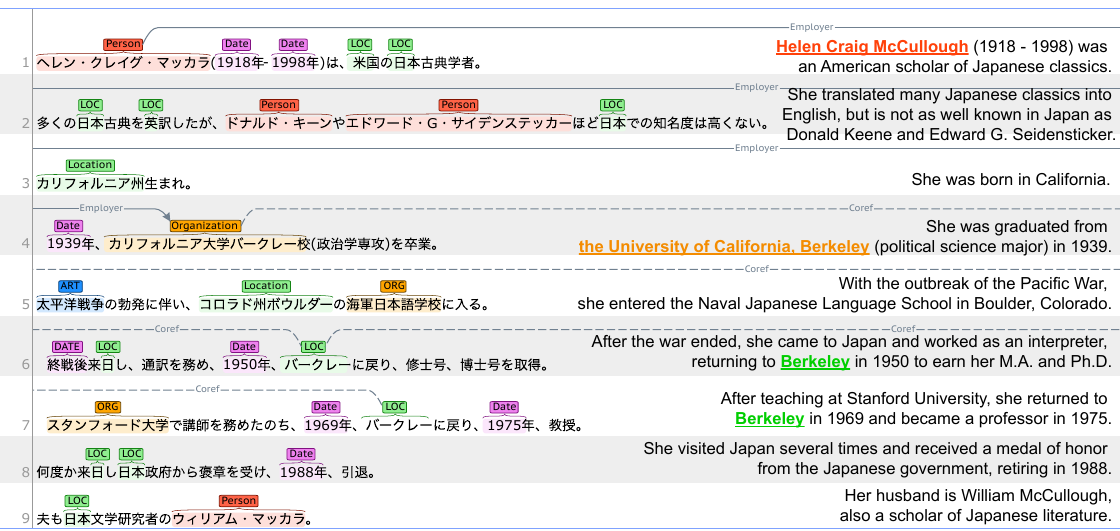}
    \caption{Interface for relation annotation. English translations are provided on the right for reference. In this example, the annotator decides whether (Helen Craig McCullough, Employer, the University of California, Berkeley) holds or not. Entity mentions connected with \textit{Coref} are coreferences of each other.}
    \label{fig:brat_example}
\end{figure*}
Relations and coreferences are annotated based on entities.
Our approach differs from existing works in that (1) we define a smaller relation label set that covers a sufficient number of relation instances, and (2) we provide machine recommendations with a model trained with Re-DocRED\textsuperscript{ja}.

\paragraph{Coreference Recommendations.}
For each entity $e_i$, we treat all its mentions $\{m_1^i, \dots, m_l^i\}$ as coreferences of each other.
As introduced in the task definition, we only consider proper nouns as mentions while excluding the pronouns.
Mentions linked to the same Wikidata entity are recommended as coreferences.

\paragraph{Relation Types.}
(Re-)DocRED's relation label set $\mathcal{R}$ contains 96 relation types.
However, it is hard for annotators to comprehend such a large label set, which will eventually affect the annotation quality.
We thus reduce the relation label set based on the following principles:
(1) All relation categories defined in ERE~\cite{song-etal-2015-light} should be covered;
(2) Explicitly-defined inverse relation pairs, e.g.,  \textit{has\_part} and \textit{part\_of}, are merged into one;
(3) Relations frequently appearing in Re-DocRED are preserved as much as possible.
This results in a label set $\mathcal{R}'$ of 28 relations covering over 88\% relation instances in Re-DocRED.

\paragraph{Relation Recommendations.}
We project the label set of Japanese Re-DocRED from $\mathcal{R}$ to $\mathcal{R}'$ and retrain a DREEAM model.
Predictions of the model are employed as machine-recommended relations.
We expect our recommendations to be more accurate than those in previous works obtained from knowledge base queries, primarily due to two factors: (1) Wikidata only stores a limited number of relation facts, while a model can, in principle, assign relation(s) to each entity pair in the document; (2) Relation facts in Wikidata are independent of the document's content, while model predictions are contextually sensitive.
A quantitative comparison of recommendations from the model trained on Re-DocRED\textsuperscript{ja} and those from querying Wikidata can be found in Section~\ref{sec:edits}.

\paragraph{Human Edits.}
Coreferences and relations are revised during human annotation.
For coreferences, human annotators remove irrelevant mentions and supplement missed mentions for each entity.
For relations, human annotators first examine the existence of each recommended relation.
As showcased in Figure~\ref{fig:brat_example}, a pair of mentions $m^h_i, m^t_i$, representing entity $e_h, e_t$ respectively, along with their relation $r$ is shown in the interface.
If annotators consider relation triple $(e_h, r, e_t)$ as true, they need to provide the evidence sentence $\mathcal{V}_{e_{h},r,e_{t}}$ within the document\footnote{Sentences where mention $m^h_i$ and $m^t_i$ resides are treated as evidence by default. Only evidence sentences other than those need to be provided.};
Otherwise, the triple should be deleted from the dataset.
Finally, the annotators supply missing relation triples and evidence sentences with their best effort.

\paragraph{Post-processing.} 
Among all 28 relation types, 7 have inverse relations defined in Wikidata.
We automatically augment triples of inversed relation types after human edits.
For example, if triple $(e_h, part\_of, e_t)$ is present in the revised annotation and relation type $part\_of$ is an inversion of $has\_part$, a new triple $(e_t, has\_part, e_h)$ will be automatically added into the annotation.
JacRED thus includes 35 relation types eventually.
A detailed list of relation types is provided in Table~\ref{tab:rel_types} of Appendix B.

\section{Dataset Analysis}

This section reports the analysis results of JacRED to provide a deeper understanding of the collected dataset.
Firstly, we compare the statistics of JacRED against (Re-)DocRED.
The comparison suggests that JacRED combines the advantages of DocRED and Re-DocRED (Section~\ref{sec:stats}).
Next, we calculate the number of edits human annotators made before reaching the final annotations.
We observe that significantly more edit steps would be necessary if the human annotation started from machine recommendations suggested by knowledge base queries (Section~\ref{sec:edits}).

\subsection{Detailed Statistics}
\label{sec:stats}

\begin{table}[t]
    \centering
    \small
    \begin{tabular}{lrrr}
    \Xhline{3\arrayrulewidth}
     & \textbf{DocRED} & \textbf{Re-DocRED} &  \textbf{JacRED} \\
     \Xhline{2\arrayrulewidth}
     \# Sentences & 7.98 & 7.98 & 8.39 \\
     \# Entities & 19.51  & 19.45 & 17.87 \\
     \# Relations & 12.45 & 29.77 & 21.12 \\
     \# Evidences & 1.60 &  0.88 & 1.67 \\ 
     \Xhline{3\arrayrulewidth}
    \end{tabular}
    \caption{Comparison of (Re-)DocRED and JacRED. Values are average for each document. }
    \label{tab:stats_detail}
\end{table}

\begin{table*}[t]
    \centering
    \small
    \begin{tabular}{lrrrr}
    \Xhline{3\arrayrulewidth}
         &  \textbf{\# Recommendations} & \textbf{\# Deletions } & \textbf{\# Substitutions } & \textbf{\# Supplements} \\
    \Xhline{2\arrayrulewidth}   
        Cross-lingual Transfer & 6,500 & 1,266 & 224 & 2,740 \\
       Knowledge Base Queries & 3,200 & 1,459 & 113 & 6,233 \\
    \Xhline{3\arrayrulewidth}
    \end{tabular}
    \caption{Number of relation instances automatically recommended and how they should be revised to reach the final human annotations.  }
    \label{tab:edits}
\end{table*}

Table~\ref{tab:stats_detail} details the agreements and differences between (Re-)DocRED and JacRED.

\paragraph{Document Complexity.}
As for document length, JacRED shares a similar scale with (Re-)DocRED at both token and sentence levels.
On one hand, documents in JacRED contain more relation instances than DocRED on average, implying that the false negative issue is mitigated in JacRED compared to DocRED.
On the other hand, documents in JacRED contain fewer relation instances than in Re-DocRED.
One possible reason is our re-definition of relation types, where symmetric relation types are merged into one as they represent the same knowledge.

\paragraph{Evidence Annotation.}
Re-DocRED revises DocRED to alleviate the false negative issue by supplying missed relation instances.
However, evidence sentences for those supplied instances are not included in Re-DocRED.
In contrast, we collect human-annotated evidence sentences during the relation annotation phase.
JacRED thus better portrays the correlation between relation and evidence sentences than Re-DocRED.

\subsection{Number of Human Edits}
\label{sec:edits}

\begin{table*}[t]
    \centering
    \small
    \begin{tabular}{lrrrrrr}
    \Xhline{3\arrayrulewidth}
    & \multicolumn{3}{c}{\textbf{Dev Set}} & \multicolumn{3}{c}{\textbf{Test Set}} \\
    \cmidrule(r){2-4} \cmidrule(r){5-7}
    & \multicolumn{1}{c}{\textbf{Rel F1}} & \multicolumn{1}{c}{\textbf{Rel F1 Ign}} & \multicolumn{1}{c}{\textbf{Evi F1}} & \multicolumn{1}{c}{\textbf{Rel F1}} & \multicolumn{1}{c}{\textbf{Rel F1 Ign}} & \multicolumn{1}{c}{\textbf{Evi F1}} \\
    \Xhline{2\arrayrulewidth}
    ATLOP~\cite{zhou2021atlop} & 66.53 & 65.21 & -- & 68.04 & 66.80 & -- \\
    DocuNet~\cite{zhang-etal-2021-document} & 66.67 & 65.37 & -- & 67.66 & 66.47 & -- \\
    KD-DocRE~\cite{tan-etal-2022-document} & 67.12 & 65.70 & -- & 68.29 & 66.99 & -- \\
    DREEAM~\cite{ma-etal-2023-dreeam} & \textbf{67.34} & \textbf{65.90} & \textbf{61.46} & \textbf{68.73} & \textbf{67.40} & \textbf{62.11} \\
    \hline
    \textit{gpt-3.5-turbo-instruct}~\cite{ouyang2022training} &  13.46  & 12.84 & -- & 13.17 & 12.90 & -- \\
    \textit{gpt-4}~\cite{openai2023gpt4} & 24.17 & 23.63 & -- & 27.45 & 26.99 & -- \\
    \Xhline{3\arrayrulewidth}
    \end{tabular}
    \caption{Models' performance on the development and test set of JacRED, with best scores \textbf{bolded}.  }
    \label{tab:mono_results}
\end{table*}

We quantify the distance between machine recommendations and human annotations of relation instances.
To this end, we compare machine recommendations against final human annotations to see how many edits have been made.
Specifically, we randomly sample 400 documents from JacRED and calculate the number of recommendations being deleted/substituted/supplied as in Table~\ref{tab:edits}. 
\paragraph{Human Annotations v.s. Machine Recommendations.}
We observe that more than 20\% of machine recommendations (1,490 out of 6,500) were regarded as inappropriate, whose relation labels were deleted or substituted by human annotators.
The human annotators also supplied another 2,740 relation instances, taking up more than 40\% of the recommendations.
We thus conclude that DocRE models trained on the automatically constructed dataset still lag behind human performance considerably, suggesting the importance of a manually collected dataset.

\paragraph{Cross-Lingual Transfer v.s. Knowledge Base Queries.}
We further measure the distance of human annotations from relations recommended by querying Wikidata, a de-facto method used in previous works~\cite{yao-etal-2019-docred,cheng-etal-2021-hacred}.
Compared with model predictions, Wikidata provides only half as many recommendations: To reach the human annotations, 50\% (1,572 out of 3,200) of the recommendations need to be revised, with another 200\% instances to be added.
In total, it takes 7,805 steps to align Wikidata recommendations with the final annotation, while only 4,230 steps are needed when employing cross-lingual transfer.
These statistics reveal the usefulness of Re-DocRED\textsuperscript{ja} in reducing human efforts.

\section{Experiments}

\paragraph{Purposes.}
We employ JacRED as a benchmark to examine the capability of existing DocRE models.
Our major concerns are:
(1) How well can existing DocRE models perform Japanese DocRE?
(2) How different can a DocRE model perform when trained on Re-DocRED\textsuperscript{ja} and JacRED?
Additionally, we evaluate the cross-lingual transferability of existing DocRE models with JacRED.
\paragraph{Settings.}
We split JacRED into train/dev/test sets with 1400/300/300 documents.
Models are trained and evaluated on a single Tesla V100 16GB GPU.
For evaluation, we follow previous works to compute the micro-averaged F1 scores for relations and evidence sentences~\cite{yao-etal-2019-docred}.
Additionally, we compute \textbf{Rel F1 Ign}, a variant of F1, where relation instances seen in the training set are ignored during evaluation.
Average scores of 5 runs initialized with different random seeds are reported throughout this paper.

\subsection{Models Trained on JacRED}
\label{sec:exp_mono}

We measure the performance of existing models when supervised by the training split of JacRED.
Specifically, we train and evaluate 4 popular models on top of \textit{tohoku-nlp/bert-base-japanese-v2} available on Huggingface\footnote{\url{https://huggingface.co/tohoku-nlp/bert-base-japanese-v2}}, with results summarized in Table~\ref{tab:mono_results}.
Among these models, DREEAM is the current state-of-the-art model on (Re-)DocRED for extracting both relations and evidence sentences.
We also evaluate the performance of LLM with in-context learning.

\paragraph{JacRED introduces extra challenges beyond those in Re-DocRED.}
In Table~\ref{tab:mono_results}, all DocRE models score above 60 on Relation F1. 
Although acceptable, the performance of each model is worse than their equivalents trained on Re-DocRED, with a gap of 10 F1 points (cf. Table~\ref{tab:mdocre}).
The result suggests potential challenges in JacRED that are absent from Re-DocRED, possibly due to the characteristics of the Japanese language, such as the omission of subjects.
Addressing such characteristics may be essential to better tackle Japanese DocRE.

\paragraph{In-context learning of LLMs on JacRED is non-trivial.}

Apart from models specially designed for DocRE, we also evaluate how LLMs can tackle the task via in-context learning.
Specifically, we pre-define the relation label set and include a pair of (document, relations) in the prompt to guide the LLM in conducting DocRE.
In our experiments, we utilize models provided by OpenAI, namely (1) the instructed version of GPT-3.5 accessed via the API key \textit{gpt-3.5-turbo-instruct}~\cite{ouyang2022training} and (2) GPT-4 accessed via the API key \textit{gpt-4}\footnote{Details of the prompt is provided in Figure~\ref{fig:prompt} of Appendix C.}.
As in the last two rows of Table~\ref{tab:mono_results}, GPT-3.5 exhibited much lower performance than the DocRE models.
GPT-4 improved over GPT-3.5 but still lagged behind the supervised DocRE models.
Similar insights have been provided by~\citet{wadhwa-etal-2023-revisiting}, where in-context learning of DocRE could not be conducted due to the length restriction of the prompt.
We succeeded in instructing LLM to conduct DocRE, while the performance is limited.
The experiment results thus highlight the challenge of DocRE as a task that LLMs cannot easily tackle.

\subsection{Models Trained on Transferred Re-DocRED}
\label{sec:exp_transfer}

\begin{table}[t]
    \small
    \centering
    \begin{tabular}{lrrr}
    \Xhline{3\arrayrulewidth}
    & \multicolumn{3}{c}{\textbf{Relation}} \\
    \cmidrule(r){2-4}
         \textbf{Training Data} & \multicolumn{1}{c}{\textbf{P}} & \multicolumn{1}{c}{\textbf{R}} & \multicolumn{1}{c}{\textbf{F1}}\\
    \Xhline{2\arrayrulewidth}
    JacRED (1,400) & \textbf{64.76} & \textbf{73.29} & \textbf{68.73} \\
    Re-DocRED\textsuperscript{ja} (3,053)  & 56.14 & 53.67 & 54.87 \\
    Re-DocRED\textsuperscript{ja} (1,400)  & 55.52 & 51.77 & 53.56 \\
    \Xhline{3\arrayrulewidth}
    \end{tabular}
    \caption{Precision (P), Recall (R), and F1 scores of DREEAM trained on different data, evaluated on the test set of JacRED.
    The number of documents in each set is shown in parentheses.}
    \label{tab:xlingual_transfer}
\end{table}

Section~\ref{sec:auto} has mentioned limitations in the dataset automatically constructed from cross-lingual transfer.
Specifically, we showcased how DocRE models trained on such a dataset fail to extract relation triples from raw Japanese documents.
Here, we quantify the performance gap between a model trained on the automatically constructed dataset (Re-DocRED\textsuperscript{ja}) and the human-annotated dataset with machine assistance (JacRED).
The test set of JacRED is adopted as the benchmark, with results shown in Table~\ref{tab:xlingual_transfer}.

\paragraph{Models trained on Re-DocRED\textsuperscript{ja} suffer from low recalls.}
From Table~\ref{tab:xlingual_transfer}, we witness that DREEAM trained on Re-DocRED\textsuperscript{ja} underperforms its equivalent trained on JacRED.
Taking a closer look at the scores, we find the gap in recalls (73.29 v.s. 53.67) is more significant than that in precisions (64.76 v.s. 56.14).
The result corresponds to our observation in Section~\ref{sec:auto} that models trained on the transferred dataset cannot identify some relation instances due to the limitation of texts translated from English.

\paragraph{The gap between models trained on Re-DocRED\textsuperscript{ja} and JacRED is evident under the same setting.}
We further train DREEAM on Re-DocRED\textsuperscript{ja} with 1,400 documents, aligned with the number of documents in JacRED.
The F1 score drops from 54.87 to 53.56, lagging behind that of the model trained on JacRED with a gap of 15 F1 points.
The results indicate that JacRED provides better supervision than Re-DocRED\textsuperscript{ja}.

\subsection{Crosslingual DocRE}
\label{sec:exp_xlingual}

JacRED also enables the evaluation of cross-lingual DocRE.
Although DocRE datasets have been collected in Chinese~\cite{cheng-etal-2021-hacred} and Korean~\cite{yang-etal-2023-histred}, they lay in different domains than (Re-)DocRED.
In contrast, JacRED is collected from Wikipedia following a pipeline similar to DocRED.
The domain and label sets of JacRED and (Re-)DocRED thus match each other, enabling the evaluation of cross-lingual DocRE.
Here, we take the first attempt to measure the cross-lingual transferability of existing models using Re-DocRED and JacRED.

\begin{table}[t]
    \centering
    \small
    \begin{tabular}{lrrrr}
    \Xhline{3\arrayrulewidth}
    & \multicolumn{3}{c}{\textbf{Rel (\textit{tgt})}} & \multicolumn{1}{c}{\textbf{Rel (\textit{src})}} \\
    \cmidrule(r){2-4}
    \textbf{Model} & \multicolumn{1}{c}{\textbf{P}} & \multicolumn{1}{c}{\textbf{R}} & \multicolumn{1}{c}{\textbf{F1}} &  \multicolumn{1}{c}{\textbf{F1}} \\
    \Xhline{2\arrayrulewidth}
    \multicolumn{4}{l}{\textbf{(a) \textit{en.} $\rightarrow$ \textit{ja.}}} \\
    ATLOP & \textbf{60.59} & 31.91 &  41.76 & 74.82 \\
    DocuNet & 60.44 & 34.50 & 43.92 & 75.02 \\
    KD-DocRE & 58.83 & \textbf{36.67} & 45.14 & 75.72 \\
    DREEAM & 60.07 & 36.36 & \textbf{45.29} & \textbf{77.22} \\
    \hline
    \multicolumn{4}{l}{\textbf{(b) \textit{ja.} $\rightarrow$ \textit{en.}}} \\  
    ATLOP & 53.13 & 48.70 & 50.72  & 64.25 \\
    DocuNet & 52.69 & 45.85 & 49.03 & 64.64 \\
    KD-DocRE & \textbf{54.22} & 50.12 & 52.09 & 65.42  \\
    DREEAM & 51.88 & \textbf{53.05} & \textbf{52.45} & \textbf{65.90} \\
    \Xhline{3\arrayrulewidth}
    \end{tabular}
    \caption{Cross-lingual performance on the test set of JacRED (\textit{ja.}) and Re-DocRED (\textit{en.}) of models with mBERT as the encoder. }
    \label{tab:mdocre}
\end{table}

Specifically, we train models on the training set in one language and evaluate them on the test set in another.
The relation label set of Re-DocRED is projected onto JacRED using the same method as in Section~\ref{sec:manual}.
To ensure the multilingualism of trained models, we adopt multilingual BERT (mBERT,~\citet{devlin-etal-2019-bert}) as the encoder.
Evaluation results are shown in Table~\ref{tab:mdocre}.

\paragraph{Cross-lingual performance of existing models is limited.}
All models exhibited a decreased accuracy in the target language.
Different from sentence-level tasks, DocRE requires not only an understanding of individual sentences but also inter-sentence semantics within the whole document, which improves the difficulty of building cross-lingual models.
This may offer a potential explanation as to why translation-based cross-lingual transfer is ineffective for DocRE, despite its successful application in sentence-level RE and OpenIE~\cite{kolluru-etal-2022-alignment,hennig-etal-2023-multitacred}.


\section{Related Work}

\paragraph{DocRE corpora in English.} 
The most well-known definition of DocRE was proposed by~\citet{yao-etal-2019-docred}, along with a dataset collected from English Wikipedia named DocRED.
While two document-level relation extraction datasets, namely CDR~\cite{li2016CDR} and GDA~\cite{wu2019GDA}, have been proposed ahead of DocRED, they were collected in the biomedical domain, thus unsuitable for developing general-purpose DocRE models.
DocRED suffers from the false negative issue where a considerable amount of relation instances are absent from the ground-truth annotations~\cite{huang-etal-2022-recommend,xie-etal-2022-eider,tan-etal-2022-revisiting}.
\citet{huang-etal-2022-recommend} randomly selected 96 documents from DocRED and relabeled them from scratch, while \citet{tan-etal-2022-revisiting} revised the whole dataset as Re-DocRED with machine assistance.
This work follows a machine-assisted annotation process as DocRED and Re-DocRED while paying extra attention to providing better machine recommendations with the model trained on a dataset transferred from Re-DocRED.

\paragraph{DocRE corpora in other languages.}
\citet{cheng-etal-2021-hacred} constructed HacRED from Chinese DBpedia to promote relation extraction from complex contexts.
\citet{yang-etal-2023-histred} focused on Korean historical RE research and collected HistRED from a travel diary written between the 16th and 19th centuries.
These datasets were collected independently from DocRED with distinct domains and label sets.
Apart from these studies, \citet{cheng-etal-2022-jamie} released a system for medical relation extraction on Japanese documents, while the dataset is not publicly available.
In this work, we explore how existing resources can help construct DocRE resources in other languages.
We share the insights that models trained under cross-lingual transfer techniques are not ready for practical use.
However, they serve as good assistants for aiding human annotations.

\paragraph{Cross-lingual transfer for structured predictions.}
Several works have adopted translation-based cross-lingual transfer approaches to solve cross-lingual and multi-lingual structured prediction tasks~\cite{faruqui-kumar-2015-multilingual,kolluru-etal-2022-alignment}.
More recently, \citet{hennig-etal-2023-multitacred} constructed MultiTACRED, a multilingual version of TACRED~\citelanguageresource{tacred}, using similar approaches as ours.
They confirmed the dataset's quality to be high enough even without human modifications.
Our work examines the approach's usefulness in the literature of DocRE and reports its shortcomings. 
Unlike other sentence-level IE tasks, DocRE involves understanding not only single sentences but also the whole document, improving the difficulty of cross-lingual transfer.

\section{Conclusion}

This work publishes JacRED, the first benchmark for general-purpose Japanese DocRE.
In the process of building JacRED, we explore how to utilize existing English DocRE resources to construct resources for other languages, using Japanese as the representative. 
Starting from constructing a dataset by translation-based cross-lingual transfer, we have shown how and why such a dataset is not ready for practical use.
Nevertheless, models trained on the dataset can replace existing approaches, i.e., knowledge base queries, to provide better recommendations for human annotation.
Our insights can benefit the development of DocRE resources for other languages.
Benchmarking with JacRED portrays the challenge of not only Japanese but also cross-lingual DocRE.

In the future, we plan to utilize models trained on JacRED to help downstream tasks such as question answering and reading comprehension.

\section{Ethics Statement}

In this work, we collected a dataset from Wikipedia, whose text content can be used under the terms of the CC-BY-SA\footnote{\url{https://en.wikipedia.org/wiki/Wikipedia:Text_of_the_Creative_Commons_Attribution-ShareAlike_4.0_International_License}}.
We thus presume that no copyright issues are involved in constructing and publishing our dataset.
\paragraph{Automatic Annotations.}

For the machine translator, we adopted DeepL API at the cost of 2,500 JPY (approx. 16\$) per 1 million characters.
For the LLM, we tested with the instructed version of GPT-3.5 provided by OpenAI at the cost of \$0.0015 per 1 thousand tokens for input and \$0.002 per 1 thousand tokens for output.
For existing DocRE models, all resources we adopted are publicly available and free of charge.

\paragraph{Human Annotations.}
Before the annotation, we arranged meetings in advance to (1) explain the purpose of collecting the dataset and (2) adjust the workload.
The annotators understand and agree that their work will be used to train neural networks.
For both the entity and relation annotation phases, we explained the purpose of building the dataset and provided a detailed annotation guideline.
During the annotation, we frequently discussed with the annotators how to handle irregular cases and adjust the guidelines when necessary.
7 annotators are involved in the entity mention annotation phase, and 6 annotators are involved in the relation annotation phase.
Each annotator is paid 5,000 JPY (approx. 30\$) per hour, which is higher than the standard salary in Japan.

\section{Acknowledgements}

This paper is based on results obtained from a project, JPNP18002, commissioned by the New Energy and Industrial Technology Development Organization (NEDO).
\nocite{*}
\section{Bibliographical References}\label{sec:reference}

\bibliographystyle{lrec-coling2024-natbib}
\bibliography{lrec-coling2024-example}

\section{Language Resource References}
\label{lr:ref}
\bibliographystylelanguageresource{lrec-coling2024-natbib}
\bibliographylanguageresource{languageresource}

\newpage

\section*{Appendix A: Entity Label Types}
\label{appendix:ent_labels}

In this section, we list all entity types in JacRED as in Table~\ref{tab:ent_types}, together with those defined in (Re-)DocRED.

\begin{table}[t]
    \centering
    \small
    \begin{tabular}{cc}
    \Xhline{3\arrayrulewidth}
        \textbf{(Re-)DocRED (6)} & \textbf{JacRED (8)}   \\
    \Xhline{2\arrayrulewidth}
        PERSON & PERSON \\
        ORGANIZATION & ORGANIZATION \\
        LOCATION & LOCATION \\
        TIME & ARTIFACT \\
        NUM & TIME \\
        MISC & DATE \\
        & PERCENT \\
        & MONEY \\
    \Xhline{3\arrayrulewidth}
    \end{tabular}
    \caption{Comparison of entity types of existing dataset and our proposed dataset. The total number of entity types is indicated in the parenthesis following each dataset.}
    \label{tab:ent_types}
\end{table}

\section*{Appendix B: Relation Label Types}
\label{appendix:rel_labels}

In this section, we list all relation types included in JacRED as in Table~\ref{tab:rel_types}.

\begin{table}[t]
    \centering
    \small
    \begin{tabular}{llr}
    \Xhline{3\arrayrulewidth}
        \textbf{ERE Category} & \textbf{JacRED Type} & \textbf{ID}   \\
    \Xhline{2\arrayrulewidth}
    Physical & Capital & P36\\
    & CapitalOf & P1376  \\
    & AdministrativeLocation & P131 \\
    & Location & P276 \\
    & WorkLocation & P937 \\
    \hline
    General & CountryOfCitizenship & P27 \\ 
    Affiliation & DateOfBirth & P569 \\
    & DateOfDeath & P570 \\
    & PlaceOfBirth & P19 \\
    & PlaceOfDeath & P20 \\
    & Follows & P155 \\
    & FollowedBy & P156 \\
    \hline
    Personal-Social & Child & P40 \\
    & Sibling & P3373 \\
    & Spouse & P26 \\
    & ParticipantIn & P1344 \\
    & Participant & P710 \\
    \hline
    Part-Whole & MemberOf & P463 \\
        & HasPart & P527 \\
        & PartsOf & P361 \\
    \hline
    Organization & HeadOfGovernment & P6 \\
    Affiliation & OwnedBy & P127 \\
    & OwnerOf & P1830 \\
    & FoundedBy & P112 \\
    & Employer & P108 \\
    & Operator & P137 \\
    & ItemOperated & P121 \\
    & EducatedAt & P69 \\
    \hline
    Others (*) & AwardReceived & P166 \\
    & Creator & P170  \\
    & Performer & P175 \\
    & Published & P123 \\
    & PresentInWork & P1441 \\
    & Characters & P674 \\
    & Platform & P400 \\
    \Xhline{3\arrayrulewidth}
    \end{tabular}
    \caption{Relation types included in our proposed dataset. Column \textbf{ID} shows the Wikidata property ID linked to each relation type. The last category \textbf{Others} includes relation types undefined in ERE type.}
    \label{tab:rel_types}
\end{table}

\section*{Appendix C: Prompt for In-Context Learning}
\label{appendix:prompt}

\begin{figure}
    \centering
    \includegraphics[width=.5\textwidth]{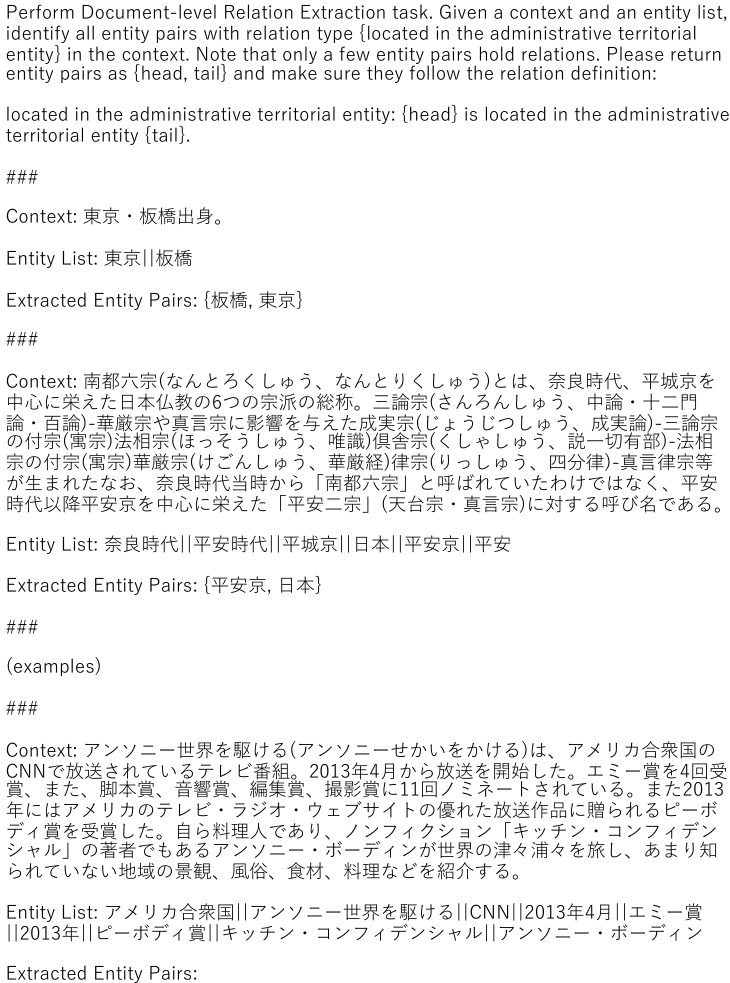}
    \caption{An example of the prompt used for the in-context learning of GPT-3.5 and GPT-4.}
    \label{fig:prompt}
\end{figure}

We showcase the prompt used for the in-context learning of LLM in Figure~\ref{fig:prompt}.
In previous work where LLM are utilized for relation extraction~\cite{wadhwa-etal-2023-revisiting,li-etal-2023-semi}, the prompt has been designed to return all relation triples within a document.
However, it is hard to identify all relation triples across a document at once.
Furthermore, most supervised approaches tackle DocRE by classifying relation types entity-pair wise~\cite{zhou2021atlop,xie-etal-2022-eider,tan-etal-2022-document,ma-etal-2023-dreeam}. 
We thus design prompts to reduce the task complexity by querying one relation type for each API call\footnote{In early experiments, we evaluated the performance when querying: 1) one relation type; 2) all relation types of one entity pair; 3) one relation type of one entity pair during each API call, where 2) yields good performance at a low cost.}.
By using our prompt, GPT-3.5 yields better performance than reported in existing works.

\end{document}